\documentclass[10pt,journal,compsoc]{IEEEtran}

\ifCLASSOPTIONcompsoc
  \usepackage[nocompress]{cite}
\else
  \usepackage{cite}
\fi

\ifCLASSINFOpdf
\else
\fi

\usepackage{times}
\usepackage{soul}
\usepackage{url}
\usepackage[hidelinks]{hyperref}
\usepackage[utf8]{inputenc}
\usepackage[small]{caption}
\usepackage{graphicx}
\usepackage{amsmath}
\usepackage{cuted}
\usepackage{widetext}
\usepackage{flushend}
\usepackage{etoolbox}
\usepackage{amsthm}
\usepackage{booktabs}
\usepackage{algorithm2e}

\AfterEndEnvironment{strip}{\leavevmode}

\makeatletter
\renewcommand{\@algocf@capt@plain}{above}
\makeatother
\SetKwInOut{Input}{Input}
\SetKwInOut{Output}{Output\,}
\SetKwInOut{Data}{Data}
\SetKwProg{ReduceSample}{ReduceSample}{}{EndReduceSample}

\usepackage[normalem]{ulem}
\usepackage{color}
\newcommand\red{\bgroup\markoverwith{\textcolor{red}{\rule[0.5ex]{2pt}{0.4pt}}}\ULon}

\usepackage{amsmath}
\usepackage{amsfonts}

\DeclareMathOperator*{\argmin}{arg\,min}

\begin{document}

\title{On the Complexity of Labeled Datasets}

\author{Rodrigo Fernandes de Mello
    \IEEEcompsocitemizethanks{\IEEEcompsocthanksitem R. F. de Mello is Chief Data Scientist at Itaú Unibanco SA, São Paulo, Brazil.\protect\\
    E-mail: rodrigo.fernandes-mello@itau-unibanco.com.br}
    \thanks{Manuscript revised June 19, 2021.}}
    

\maketitle

\begin{abstract}
The Statistical Learning Theory (SLT) provides the foundation to ensure that a supervised algorithm generalizes the mapping $f: \mathcal{X} \to \mathcal{Y}$ given $f$ is selected from its search space bias $\mathcal{F}$. SLT depends on the Shattering coefficient function $\mathcal{N}(\mathcal{F},n)$ to upper bound the empirical risk minimization principle, from which one can estimate the necessary training sample size to ensure the probabilistic learning convergence and, most importantly, the characterization of the capacity of $\mathcal{F}$, including its underfitting and overfitting abilities while addressing specific target problems. However, the analytical solution of the Shattering coefficient is still an open problem since the first studies by Vapnik and Chervonenkis in $1962$, which we address on specific datasets, in this paper, by employing equivalence relations from Topology, data separability results by Har-Peled and Jones, and combinatorics. Our approach computes the Shattering coefficient for both binary and multi-class datasets, leading to the following additional contributions: (i) the estimation of the required number of hyperplanes in the worst and best-case classification scenarios and the respective $\Omega$ and $O$ complexities; (ii) the estimation of the training sample sizes required to ensure supervised learning; and (iii) the comparison of dataset embeddings, once they (re)organize samples into some new space configuration. All results introduced and discussed along this paper are supported by the R package shattering~\url{https://cran.r-project.org/web/packages/shattering}.
\end{abstract}
\vspace{1cm}
\IEEEpeerreviewmaketitle
\IEEEraisesectionheading{\section{Introduction}\label{sec:introduction}}

\IEEEPARstart{T}{he} Statistical Learning Theory (SLT) is amongst the most important results for supervised machine learning~\cite{Vapnik2013nature}. SLT formalizes the Empirical Risk Minimization Principle (ERMP) which ensures the probabilistic convergence of the empirical risk $R_\text{emp}(f) \in [0,1]$ to its expected value, a.k.a. risk $R(f) \in [0,1]$, for some classification function $f$. By having access to the risk of functions $f_1, \ldots, f_m$, one can select the best mapping $f:\mathcal{X} \to \mathcal{Y}$ from $f_\mathcal{F} = \argmin R(f_i)$, for $i = 1, \ldots, m$, by assuming that $R(f)$ is computed on the Joint Probability Distribution (JPD) $P(\mathcal{X},\mathcal{Y})$, given $\mathcal{X}$ and $\mathcal{Y}$ are the input and output spaces, respectively.

Unfortunately, the JPD is unknown for real-world problems~\cite{vonLuxburg} given the inherently continuous and unbounded nature of input attributes leading to the impossibility of computing the risk $R(f)$ for any $f$. Thus, by assuming the i.i.d.\ (independent and identically distributed) sampling from such a JPD, Vapnik~\cite{Vapnik2013nature} employed the Uniform Law of Large Numbers (ULLN) and the Chernoff's bound~\cite{Devroye96} to obtain the following result:
{\footnotesize
\begin{align}\label{eq:ermp}
    P(\sup_{f \in \mathcal{F}} |R_\text{emp}(f)-R(f)| > \epsilon) \leq 2 \mathcal{N}(\mathcal{F},n) \exp{(-n \epsilon^2/4)},
\end{align}}
in which $\mathcal{F}$ defines the subspace of admissible functions of a given supervised learning algorithm, a.k.a. the search space bias an algorithm selects classification functions from; $\epsilon \in [0,1]$ refers to an acceptable divergence between risks $R_\text{emp}(f)$ and $R(f)$; $n$ is the sample size; and $\mathcal{N}(\mathcal{F},n)$ corresponds to the Shattering coefficient or the growth function mapping the number of distinct classifications an algorithm outputs as sample size $n$ increases.

As later proved by Sauer~\cite{Sauer72jct} and Shelah~\cite{CIS-393795}, the Shattering coefficient function is upper bounded as defined in Inequation~\ref{eq:sc}, in which $\text{VC}$ is the Vapnik-Chervonenkis dimension for a target scenario, corresponding to the greatest sample that can be divided/separated/classified in all $2^\text{VC}$ possibilities provided by a binary space segmentation. For instance, consider one point in $\mathbb{R}^2$ and a single hyperplane, we can classify it as either positive or negative by simply placing a hyperplane on both sides of such a point. Thus, using a $(p-1)$-dimensional hyperplane, the greatest sample that can be classified in all possible ways in some space $\mathbb{R}^p$ is $p+1$, so that $\text{VC}(\mathbb{R}^p) = p+1$~\cite{vonLuxburg,Vapnik2013nature}.
\begin{align}\label{eq:sc}
    \mathcal{N}(\mathcal{F},n) \leq \sum_{i=0}^{\text{VC}} \binom{n}{i},
\end{align}
Considering that the precise computation of the Shattering coefficient has been an open problem since the first studies by Vapnik and Chervonenkis in $1962$~\cite{Chervonenkis2013}, it is very complex, or even impossible, to measure the space of admissible functions of an algorithm. Therefore, after the recent theoretical contributions by Har-Peled and Jones~\cite{Har-Peled2019} in the context of dataset partitioning, we introduce in this paper a tighter estimation of the Shattering coefficient function considering binary and multi-class datasets. Furthermore, based on the same result, we also estimate the number of hyperplanes required to shatter a given sample in terms of $\Omega$ and $O$ complexities, leading to the separation of every pair of points under different classes from one another. It is important to highlight that, the number of hyperplanes is another relevant contribution of this work to help parametrizing learning algorithms, what is specially motivating in the context of deep neural networks~\cite{DBLP:journals/eswa/FerreiraCNM18}. Lastly, to complement our contributions, the R Package named ``shattering'' was developed and made available at~\url{https://cran.r-project.org/web/packages/shattering}.

This paper is organized as follows: Section~\ref{sec:contextualization} describes the motivating problem and summarize our contributions; Section~\ref{sec:background} briefly introduces the background concepts and some related work; Our approach to compute the Shattering coefficient function is detailed in Section~\ref{sec:shattering}; and concluding remarks and future directions are drawn in Section~\ref{sec:concluding}.

\section{Contextualization}\label{sec:contextualization}

In order to exemplify our contributions, consider a simple binary classification problem whose Joint Probability Distribution $P(\mathcal{X}, \mathcal{Y})$ is composed of two well-separated bidimensional Gaussian functions, as illustrated in Figure~\ref{fig:gaussians}(a). In this case, we employed Inequation~\ref{eq:sc} to estimate the Sauer-Shelah worst-case classification scenario for the Shattering coefficient function $\mathcal{N}(\mathcal{F},n)_\text{Sauer-Shelah} \leq \sum_{i=0}^{3} \binom{n}{i}$, given $\text{VC}=3$. However, observe that such a JPD is characterized by two homogeneous regions, each one used to identify a single class of a binary problem, therefore, the inclusion of points into such a well-separated space should not drastically change the number of classification possibilities a supervised learning algorithm is capable of representing. Thus, we derived lower and upper bounds for the Shattering coefficient function, as detailed in Section~\ref{sec:shattering}.

As a result, we conclude that $\mathcal{N}(\mathcal{F},n)_\text{Sauer-Shelah}$ is a very far upper bound for the Shattering coefficient function, while our approach allows to better estimate the capacity of the space of admissible functions $\mathcal{F}$, consequently impacting on:
\begin{itemize}
    \item The estimation of the required number of hyperplanes in the best and worst-case scenarios using the complexity notations $\Omega$ and $O$ -- this permits the proper study, analysis and definition of learning architectures, such as the number of units in an artificial neural network;
    \item The lower and upper bound Shattering functions permit the estimation of the sample sizes required to ensure supervised learning for practical applications -- the use of those bounds on the ERMP (Inequation~\ref{eq:ermp}) is used to provide such an estimation;
    \item A tighter estimation for the Shattering coefficient function rather than the Sauer-Shelah's approach;
    \item The shattering formulation for multi-class classification problems; and
    \item The analysis of the influence of the number of hyperplanes on the Shattering coefficient function.
\end{itemize}

Furthermore, we allow the comparison of data embeddings, given each of those is responsible for reorganizing examples into some new space configuration. As a consequence, each embedding instance can be assessed using our approach to conclude on the space separability difficulty, thus leading to a embedding complexity measure. From that, different space embeddings can be compared to one another in an attempt to select the most adequate to address a given learning task.

Finally, all those contributions together allow a more precise analysis on the space of admissible functions, a.k.a. the algorithm search bias $\mathcal{F}$, as well as the bias comparison against different learning settings. 

\section{Background and Related Work}\label{sec:background}

Vapnik~\cite{Vapnik2013nature} formulated the Statistical Learning Theory (SLT) to provide upper bounds to the probabilistic convergence of the empirical risk to its expected value, given a set of classification functions in form $f_i : \mathcal{X} \to \mathcal{Y}$, thus mapping input instances from $\mathcal{X}$ to output labels in $\mathcal{Y}$. Inequation~\ref{eq:ermp} was deduced in such a context, which was later used to formalize the Generalization Bound as follows:
\begin{align}\label{eq:gbound}
    R(f) \leq R_\text{emp}(f) + \sqrt{\frac{4}{n} \left( \log{2\mathcal{N}(\mathcal{F},n)} - \log{\delta} \right)},
\end{align}
assuming $P(\sup_{f \in \mathcal{F}} |R_\text{emp}(f)-R(f)| > \epsilon) \leq \delta$. Therefore, by having the Shattering coefficient function $\mathcal{N}(\mathcal{F},n)$, one can formulate how the risk $R(f)$ is upper bounded by the empirical risk $R_\text{emp}(f)$ plus some variance. As a clear consequence, by computing this function for real-world tasks, one can understand this estimation process as well as define the minimal sample size to guarantee learning convergence as discussed in~\cite{DBLP:books/sp/MelloP18}.

To illustrate the aforementioned, consider the estimation of the necessary training sample size by assuming $\mathcal{N}(\mathcal{F},n)=n^2$. From Inequation~\ref{eq:ermp}, we have:
\begin{align*}
    \delta  = 2 n^2 \exp{(-n \epsilon^2 / 4)}
            & = 2 \exp{\log{2 n^2}- n \epsilon^2 / 4}\\
            & = 2 \exp{\log{n^2} + \log{2} - n \epsilon^2 / 4}\\
            & = 2 \exp{(2\log{n} + \log{2} - n \epsilon^2 / 4)},
\end{align*}
given an acceptable value for $\delta$ to ensure the probabilistic convergence. To do so, suppose $\delta=0.05$ and maximum acceptable divergence between risks $\epsilon=0.01$, we obtain:
\begin{align*}
    0.05 = 2 \exp{(2\log{n}  + \log{2} - n 0.01^2 / 4)},
\end{align*}
so that solving for $n \geq 1$, we obtain $n \approx 1,301,610$ as the necessary training sample size to guarantee such a probabilistic convergence, as follows:
\begin{align*}
    P(\sup_{f \in \mathcal{F}} |R_\text{emp}(f)-R(f)| > 0.01) \leq 0.05,
\end{align*}
so the empirical risk $R_\text{emp}(f)$ diverges more than $0.01$ units from the risk $R(f)$ with a probability less than or equals to $0.05$, therefore in $95\%$ of the scenarios $R_\text{emp}(f)$ is a good estimator for its expected value according to the divergence factor $\epsilon=0.01$. As main consequence, there is a significant guarantee of selecting an adequate learning model to address this task, once one knows how it operates on unseen data examples~\cite{DBLP:books/sp/MelloP18}.

Still considering Inequation~\ref{eq:gbound} for $\delta=0.05$, we find:
\begin{align*}
    R(f) \leq R_\text{emp}(f) + \sqrt{\frac{4}{n} \left( 2 \log{n} +\log{2} - \log{0.05} \right)},
\end{align*}
so one obtain the divergence between $R(f)$ and $R_\text{emp}(f)$ as the sample size $n$ increases. It is noteworthy to observe that the rightmost term approaches zero as $n \to \infty$, such that the empirical risk becomes an ideal estimator of $R(f)$.

From a different perspective, Sauer~\cite{Sauer72jct} proved that if the density of a family $\mathcal{F}$ of subsets of a set $\mathcal{S}$ with $|\mathcal{S}|=m$ is less than $n$, then:
\begin{align*}
    |\mathcal{F}| \leq \sum_{i=0}^{n-1} \binom{m}{i},
\end{align*}
there exists a family $\mathcal{F}$ of subsets of $\mathcal{S}$ with $|\mathcal{F}| = \sum_{i=0}^{n-1} \binom{m}{i}$, such that the density of $\mathcal{F}$ is $n-1$ ($m \geq n \geq 1$), allowing to find a theoretical upper bound for the Shattering coefficient function as defined in Inequation~\ref{eq:sc}. This density was then proved to be expressed by the Vapnik-Chervonenkis (VC) dimension~\cite{Vapnik2013nature}.

The aforementioned formulation of the Shattering coefficient provides an important upper bound, however it is worth highlighting: (i) it tends to be very far from the growth function itself, given it does not consider data distributions; (ii) it depends on the VC dimension which may be difficult to estimate~\cite{vonLuxburg}; (iii) it does not take into account the number of hyperplanes adopted in practical application scenarios; and (iv) it does not consider multi-class classification problems.

\section{On the supervised data complexity}\label{sec:shattering}

Har-Peled and Jones~\cite{Har-Peled2019} have recently published a fundamental theoretical result on data partitioning that supports the approach introduced in this paper to compute the Shattering coefficient function for specific datasets. Given a set $D$ with $n$ points in general position, i.e., any three-point arrangement cannot be collinear, lying inside a $d$-dimensional unit cube $[0,1]^d$ for $d \geq 1$, Har-Peled and Jones proved a theorem to compute the minimum number of hyperplanes necessary to separate -- a property referred to as separability and denoted as $\text{Sep}(D)$. They performed a demonstration and concluded that the minimum number of hyperplanes separating $D$ is $\Omega(n^\frac{2}{d+1} \log \log n / \log n)$ and, in expectation, one can separate $D$ using $O(d n^\frac{2}{d+1})$ hyperplanes.

Let some supervised dataset $D = \{(x_1,y_1),(x_2,y_2),\ldots,(x_n,y_n)\}$ be composed of pairs $x \in \mathcal{X}$ and $y \in \mathcal{Y}$ corresponding to an input example and its corresponding class label, respectively. Then, suppose the input space $\mathcal{X} \subseteq \mathbb{R}^d$ presents different degrees of class overlapping among examples in $D$ which obviously jeopardize the classification process. Using the theorem by Har-Peled and Jones~\cite{Har-Peled2019}, we firstly define some cube in $\mathbb{R}^d$ to enclose all examples and, then, we rescale their relative distances in order to ensure every space axis is in range $[0,1]$ so that a cube $[0,1]^d$ is obtained for $d \geq 1$.

Afterwards, we employ the expectation by Har-Peled and Jones~\cite{Har-Peled2019} to estimate the number of hyperplanes to separate $D$, i.e., $\text{Sep}(D)$, such that we have the minimum number of linear functions providing the pairwise separability of points. According to the SLT, this corresponds to the overfitting scenario in which every example in $D$ is separated from every other, thus tending to the memory-based classifier, as discussed in~\cite{Vapnik2013nature,vonLuxburg,DBLP:books/sp/MelloP18}. Although this represents the overfitting scenario, it also works as an upper bound for the required number of hyperplanes to shatter such an input space.

\subsection{Number of homogeneous-class regions}

\begin{figure}[!htb]
    \centering
    \includegraphics[scale=0.45]{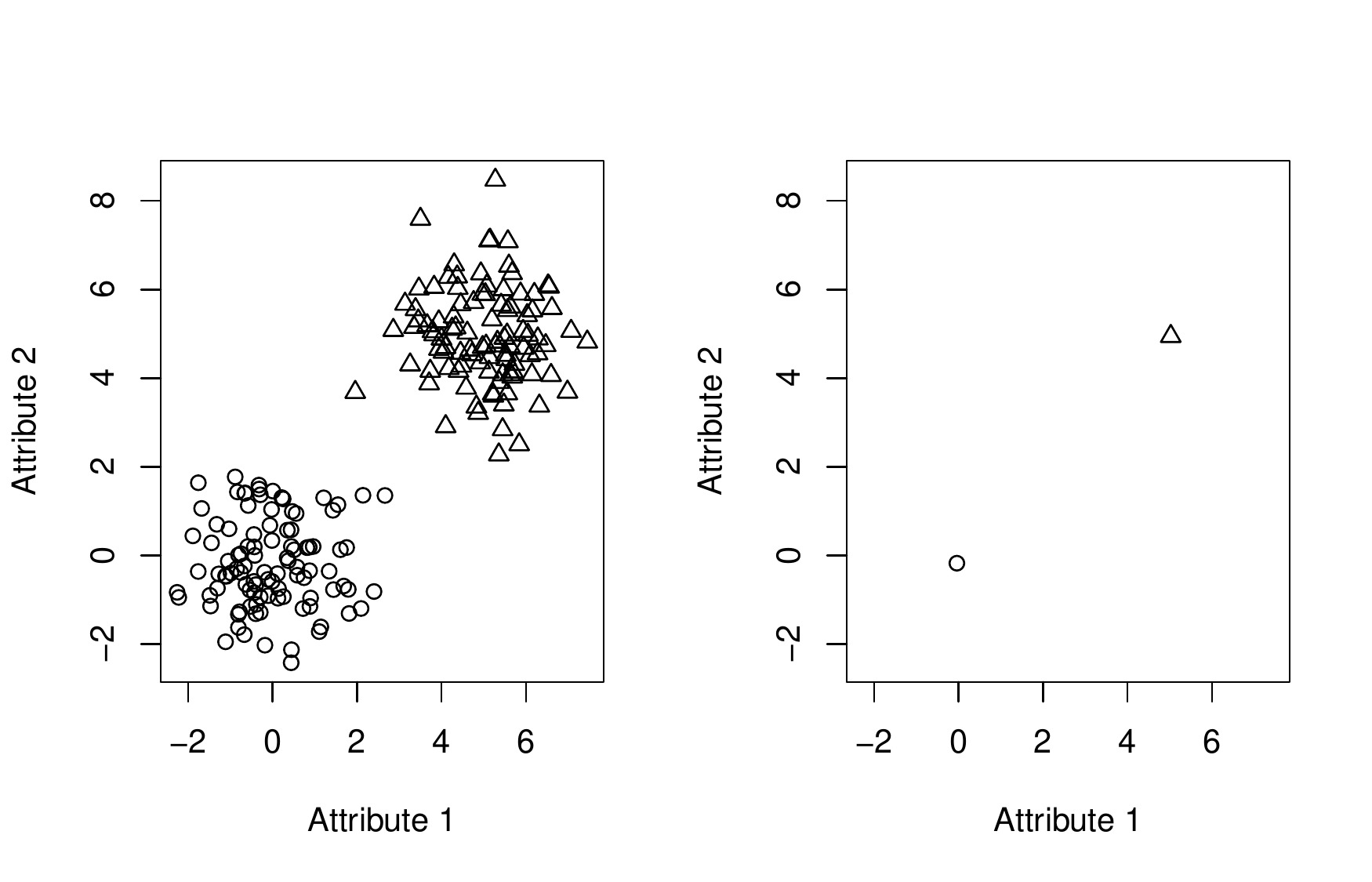}
    \caption{Input space built upon two 2D Gaussian distributions with averages $(0,0)$ and $(10,10)$, respectively. Their variances are equal to $1$ along both dimensions. From left to right: (i) the input space is illustrated using two different symbols (one per class); and (ii) the reduced space after exploring homogeneous class neighborhoods.}\label{fig:gaussians}
\end{figure}

Based on Har-Peled and Jones' results~\cite{Har-Peled2019}, we observed that the number of hyperplanes found with $O(d n^\frac{2}{d+1})$ did not correspond to the expected number when addressing practical problems. For instance, consider Figure~\ref{fig:gaussians}(a) that illustrates a two-class problem built upon two 2D Gaussian distributions parametrized with the averages $(0,0)$ and $(5,5)$, respectively, whose variances are set as one along both dimensions. Instead of analyzing the sample size $n$, leading to a growing number of hyperplanes necessary to separate each point from one another, we here observe the opportunity to measure the number of hyperplanes as the number of homogeneous-class regions grows.

To do so, we circumscribe such an input space in $[0,1]^2$ and then count the number of homogeneous-class regions as the sample size $n$ increases. Knowing that this particular instance will never overlap labels from different classes, then we know that the number of regions $r(.)=2$ is constant and equals to two. Thus, according to Har-Peled and Jones~\cite{Har-Peled2019}, we estimate the number of hyperplanes $O(d n^\frac{2}{d+1})$ as $O(2 \times 2^\frac{2}{2+1})$ for $d=2$ (input space is bidimensional) and $n=2$, once the input space is reduced to two homogeneous-class regions independently of any increase in the number of samples. This is only possible because such an input space presents a clear linear separation of instances under different class labels.

After that, we analyze the original input space using the most relevant examples, such as the support vectors in case of SVM~\cite{Cortes1995}, in an attempt to obtain some other space as illustrated in Figure~\ref{fig:gaussians}(b), which represents homogeneous-class (contiguous) regions that would be separated by the minimal number of hyperplanes as $n \to \infty$. Then, we compute the number of homogeneous-class regions using the equivalence relation $x_i ~ x_j$ in between every pair of examples in $D$, having $x_i,x_j \in \mathbb{R}^d$ and their corresponding class labels as $y_i,y_j \in \mathbb{Z}_+$, as discussed in~\cite{mendelson1990introduction}.

As next step, we build up the equivalence relations of an input example $x_i$ by measuring the closest point $x_k$ belonging to any other class than $y_i$, so that radius $\theta_i < f_d(x_i, x_k)$ for some metric distance $f_d$. Observe that this allows to devise an open-ball $B_{\theta_i}(x_i)$ around $x_i$, connecting it to any other point $x_j$ belonging to the same class, i.e., $y_i = y_j$. This is the strategy we use to build up the equivalence relationship among all input examples in $X \subset \mathcal{R}^d$.

After constructing all open balls $B_{\theta_i}(x_i)$, for all $i=1,\ldots,n$, we connect every point $x_i$ to its equivalent $x_j$ contained inside the open ball of $x_i$, so we create homogeneous-class regions along the input space. This strategy is performed as the sample size $n$ increases, thus leading to a Cartesian plane whose abscissa corresponds to the sample size $n \in \mathcal{Z}_+$ and the ordinate represents the number of homogeneous-class regions we obtain as $n \to \infty$. Then, we perform a regression $h(n)$ on top of such a plane to obtain the relation on the number of space regions as the sample size $n$ grows. Obviously, if no class overlapping occurs as the sample size increases, the number of regions is constant. 

Note that function $h(n)$ is either linear or sublinear once the number of homogeneous regions is always smaller than or equal to the sample size.

\subsection{Number of hyperplanes}

At this point, we apply the theoretical results by Har-Peled and Jones~\cite{Har-Peled2019} on function $h(n)$ to estimate the lower and the upper bounds for the number of hyperplanes required to solve such a classification problem. Thus, the number of hyperplanes for the best and worst-case scenarios $\Omega(n^\frac{2}{d+1} \log \log n / \log n)$ and $O(d n^\frac{2}{d+1})$, respectively, are rewritten as $\Omega(h(n)^\frac{2}{d+1} \log \log h(n) / \log h(n))$ and $O(d h(n)^\frac{2}{d+1})$, leading to the separability of $D$ according its homogeneous-class regions.

As a result, one can determine the range of the number of hyperplanes to address a given classification task, from the lowest to its greatest number. Observe that this result provides the guidelines to 
parametrize the number of units of a single hidden layer in a Multilayer Perceptron~\cite{DBLP:books/sp/MelloP18} or even the convolutional units of a single-layer Deep Learning approach~\cite{DBLP:journals/eswa/FerreiraCNM18}. In addition, one can compare the number of hyperplanes induced by any other algorithm, such as a Decision Tree, and compare their complexities considering that more hyperplanes lead to more complex models. Furthermore, observe that several different space embeddings could be assessed using what we introduced so far, for instance, one can have the ability of correctly determining the number of units to employ in a $3 \times 3$ convolutional mask of a Deep Learning layer.

\subsection{Computing the Shattering coefficient function}

We can now follow two different paths to measure the Shattering coefficient: either (i) by considering the maximal number of regions after applying Har-Peled and Jones' approach to find $\Omega(h(n)^\frac{2}{d+1} \log \log h(n) / \log h(n))$ and $O(d h(n)^\frac{2}{d+1})$; or (ii) by using function $h(n)$ to account for the number of space regions in $\mathbb{R}^d$.

By employing any of the strategies, we reduce the classification problem to a region-coloring problem. Suppose we have two class labels ($C=2$) and $r$ regions, the following combinatorics account for all space-coloring possibilities:
\begin{align*}
    1 + \sum_{c_1=1}^\text{r} \binom{\text{r}}{c_1} = 2^r,
\end{align*}
representing the Shattering coefficient function. For instance, consider $\mathcal{R}^2$ with two-nonparallel hyperplanes shattering such input space so four different regions would be obtained and, thus $2^4 = 8$ classification possibilities are drawn as expected. Notice this is the same as defining which space region will be colored with an alternative color other than black, for example.

By considering an additional class, i.e., for $C=3$, the following combinatorics will hold:
\begin{align*}
    1 + \sum_{c_1=1}^{r} \sum_{c_2=1}^{r-c_1} \binom{r}{c_1} \binom{r-c_1}{c_2} = 3^r -2^(r + 1) + 2,
\end{align*}
implying another sum to decide on the colors for $c_2 = 1, \ldots, r-c_1$ regions, given $c_1$ regions were already colored by the first summation term. The first binomial $\binom{r}{c_1}$ takes into account the division between the first two colors, while the second binomial $\binom{r-c_1}{c_2}$ considers the remaining regions that could be colored with a third option.

Using this rationale, we can follow with the Shattering coefficient estimation by using (i), we define the number of regions a $\mathbb{R}^d$ space can be divided using $m$ hyperplanes, as follows:
\begin{align*}
    r(m,d) = 1 + \sum_{i=1}^{d} \binom{m}{i}
\end{align*}
so the Shattering coefficient function is bounded, for $C \geq 2$ classes, given $\Omega = \Omega(h(n)^\frac{2}{d+1} \log \log h(n) / \log h(n))$ and $O = O(d h(n)^\frac{2}{d+1})$, as follows:
\begin{widetext}
{\small
\begin{gather*}
1+\sum_{c_1=1}^{r(\Omega,d)} \sum_{c_2=1}^{r(\Omega,d)-c_1} \ldots \sum_{c_{C-1}=1}^{r(\Omega,d)-c_1-c_2\ldots-c_{C-2}} \binom{r(\Omega,d)}{c_1} \times \binom{r(\Omega,d)-c_1}{c_2}\times \ldots \times \binom{r(\Omega,d)-c_1-c_2\ldots-c_{C-2}}{c_{C-1}}\\ \leq \mathcal{N}(\mathcal{F},n) \leq\\ 1+\sum_{c_1=1}^{r(O,d)} \sum_{c_2=1}^{r(O,d)-c_1} \ldots \sum_{c_{C-1}=1}^{r(O,d)-c_1-c_2\ldots-c_{C-2}} \binom{r(O,d)}{c_1} \times \binom{r(O,d)-c_1}{c_2}\times \ldots \times \binom{r(O,d)-c_1-c_2\ldots-c_{C-2}}{c_{C-1}},
\end{gather*}}
\end{widetext}
\begin{widetext}
Following the Shattering coefficient estimation with (ii), we may simply substitute $r(\Omega,d)$ and $r(O,d)$ by $\left \lceil h(n) \right \rceil$, thus allowing to compute a closer bound, in form:

\small{\begin{align*}
\mathcal{N}(\mathcal{F},n) \approx 1+\sum_{c_1=1}^{\left \lceil h(n) \right \rceil} \sum_{c_2=1}^{\left \lceil h(n) \right \rceil-c_1} \ldots \sum_{c_{C-1}=1}^{\left \lceil h(n) \right \rceil-c_1-c_2\ldots-c_{C-2}} \binom{\left \lceil h(n) \right \rceil}{c_1} \times \binom{\left \lceil h(n) \right \rceil-c_1}{c_2}\times \ldots \times \binom{\left \lceil h(n) \right \rceil-c_1-c_2\ldots-c_{C-2}}{c_{C-1}},
\end{align*}}
resulting in a tighter approximation to what the equivalence relations represent in terms of topological features. 
\end{widetext}

\subsection{On the R package shattering}

All features discussed so far were implemented and made available with the R package shattering~\footnote{The R package shattering is available at \url{https://cran.r-project.org/web/packages/shattering}}, which provides the following functions:
\begin{enumerate}
    \item equivalence\_relation -- this function builds up one open-ball per point in space $\mathbb{R}^d$, whose radius is infinitesimally smaller than the distance sufficient to connect such a point to any its closest points belonging to a different class (any other than its own), as previously discussed;  
    \item compress\_space -- this function uses the open-balls built up from the previous function to generate homogeneous-class regions, i.e., regions containing a single-class examples; 
    \item estimate\_number\_hyperplanes -- this function applies Har-Peled and Jones' approach to estimate the number of hyperplanes for the best and worst-case scenarios, as previously detailed;
    \item number\_regions -- from the number of hyperplanes and the space dimensionality, this function computes the maximal number of regions contained in such dataset instance under analysis;
    \item complexity\_analysis -- this function performs all previous functions on top of some supervised dataset to estimate the lower and the upper bounds for the Shattering coefficient function;
    \item build\_classifier -- this builds up an ideal classification function taking into account all equivalence relations among examples in the dataset, using radial Gaussian functions; 
    \item apply\_classifier -- this applies the classification function induced by build\_classifier into some test set.
\end{enumerate}

In this context, the most important function is complexity\_analysis which randomly selects examples from dataset $D = \{(x_1,y_1),(x_2,y_2),\ldots,(x_n,y_n)\}$ while increasing the sample size to compute the number of homogeneous-class regions $h(n)$. Then, it computes the number of hyperplanes using Har-Peled and Jones' approach, which is then used to account for the number of (homogeneous-class) space regions, and, finally, estimate the Shattering coefficient bounds.

In addition, function complexity\_analysis still provides an auxiliary code so the user can obtain two closed-form functions to represent the lower and upper bounds for the Shattering coefficient, respectively. As presented in the next sections, those closed-form functions are then used to: (i) measure the acceptable divergence $\epsilon$ (Inequation~\ref{eq:ermp}) in between the empirical risk $R_\text{emp}(f)$ and its expected value $R(f)$; and (ii) measure the variance term for the Generalization bound (squared-root term in Inequation~\ref{eq:gbound}). 

\subsection{From the Shattering coefficient to risks divergence}

From both closed-form functions to represent the lower and upper bounds for the Shattering coefficient, we estimate the minimal acceptable divergence $\epsilon$ in between the empirical risk $R_\text{emp}(f) \in [0,1]$ and the (expected) risk $R(f) \in [0,1]$. The important contribution indicates how close both risks are one to another and, consequently, how close the estimator $R_\text{emp}(f)$ is of $R(f)$ for the dataset under analysis. As a practical result, the closer the estimator is, the stronger are the learning guarantees.

Suppose the closed-form functions produced by our R package shattering are:
\begin{align*}
    \exp{(0.001n)} \leq \mathcal{N}(\mathcal{F},n) \leq \exp{(0.01n)},
\end{align*}
thus substituting both in the Empirical Risk Minimization Principle (ERMP), from Inequation~\ref{eq:ermp}, we have:
\small{\begin{align*}
    &P_\text{lower}(\sup_{f \in \mathcal{F}} |R_\text{emp}(f)-R(f)| > \epsilon) \leq 2 \exp{(0.001 n)} \exp{(-n \epsilon^2 / 4)}\;\\ &\text{and}\\
    &P_\text{upper}(\sup_{f \in \mathcal{F}} |R_\text{emp}(f)-R(f)| > \epsilon) \leq 2 \exp{(0.01 n)} \exp{(-n \epsilon^2 / 4)},
\end{align*}}
so that the uniform convergence only happens if the magnitude of the negative exponential term dominates the positive associated to the Shattering coefficient functions. Thus, we proceed as follows:
\small{\begin{align*}
    &\text{From:}\\
    &2 \exp{(0.001 n)} \exp{(-n \epsilon_\text{lower}^2 / 4)} = 2 \exp{(0.001 n -n \epsilon_\text{lower}^2 / 4)}\\
    &\text{we solve:}\\
    &0.001 n < n \epsilon_\text{lower}^2 / 4\\
    &\epsilon_\text{lower} \approx \sqrt{4 \times 0.001},
\end{align*}}
for the lower bound, and:
\small{\begin{align*}
    &\text{From:}\\
    &2 \exp{(0.01 n)} \exp{(-n \epsilon_\text{upper}^2 / 4)} = 2 \exp{(0.01 n -n \epsilon_\text{upper}^2 / 4)}\\
    &\text{we solve:}\\
    &0.01 n < n \epsilon_\text{upper}^2 / 4\\
                     &\epsilon_\text{upper} \approx \sqrt{4 \times 0.01},
\end{align*}}
for the upper bound of $\epsilon$ when considering both Shattering coefficient functions.

From this analysis, we have the minimal divergence $\sqrt{4 \times 0.001} \leq \epsilon \leq \sqrt{4 \times 0.01}$ for which the uniform learning convergence is ensured~\cite{Vapnik2013nature}. For example, given the empirical and the expected risks are in $[0,1]$, we wish to have the smallest as possible value for $\epsilon \to 0$. If one wishes to compare this dataset against another one devoted to solve the same classification task, $\epsilon$ is a great basis for comparison. The smaller it is, the more separable the dataset is.

The minimal training sample size to ensure learning is then resultant of the value $\epsilon + \zeta$, for any $\zeta > 0$, thus leading to:
\begin{align*}
    P(\sup_{f \in \mathcal{F}} |R_\text{emp}(f)-R(f)| > \epsilon) \leq 2 \exp{(-n \zeta^2 / 4)},    
\end{align*}
which can be solved for $n > 0$ and find the minimal sample size to train a supervised model, so assuming:
\begin{align*}
    \delta = 2 \exp{(-n \zeta^2 / 4)},    
\end{align*}
given $\delta$ is the probability you wish to ensure. 

Suppose one assumes $\delta=0.05$, so that $95\%$ of confidence is provided as learning guarantee, and set $\zeta = 0.001$, then the following will hold:
\begin{align*}
    0.05 = 2 \exp{(-n\; 0.001^2 / 4)}\\
    n \approx 1.47555 \times 10^7,
\end{align*}
but maybe you decided to set $\zeta = 0.01$, then you will need far less training examples:
\begin{align*}
    0.05 = 2 \exp{(-n\; 0.01^2 / 4)}\\
    n \approx 147,555.
\end{align*}

It is worth highlighting that this contribution allows the comparison of datasets and/or feature spaces, e.g. embeddings, in an attempt of assessing their inherent separability. The more separable such space is, the simpler will be its modeling and stronger its learning guarantees.

\subsection{On the Generalization bound}

After generating the lower and upper bounds for the Shattering coefficient, one may also study the Generalization Bound as a strategy to assess the implications of the square-root term (Inequation~\ref{eq:gbound}). Using the same closed-form functions as in the previous section, the formulation for the lower bound is:
\begin{align*}
    R_\text{lower}(f) \leq R_\text{emp}(f) + \sqrt{\frac{4}{n} \left( \log{(2 \exp{(0.001n)})} - \log{\delta} \right)},
\end{align*}
and the upper bound is:
\begin{align*}
    R_\text{upper}(f) \leq R_\text{emp}(f) + \sqrt{\frac{4}{n} \left( \log{(2 \exp{(0.01n)})}- \log{\delta} \right)},
\end{align*}
then, assuming $\delta = 0.05$ and solving for $n \to \infty$, we find:
\begin{align*}
    R_\text{lower}(f) \leq R_\text{emp}(f) + 0.0632456\\    
    R_\text{upper}(f) \leq R_\text{emp}(f) + 0.2
\end{align*}
from which we conclude there is a variance or risk perturbation around $6\%$ when using the lower-bound Shattering coefficient (best-case scenario) and $20\%$ when taking the the upper bound (worst-case scenario) into account.

From a practical perspective, suppose one assumes $R_\text{emp}(f)=0.01$, from which the accuracy is $1-R_\text{emp}(f)=0.99$, what could be seen as a great classification result. However, from the dataset perspective, our sample is not good enough to ensure such learning guarantee, simply because the variances resultant of the squared-root term from Inequation~\ref{eq:gbound} are greater than the empirical risk itself.

\section{Concluding remarks}\label{sec:concluding}

In this paper, we proposed a new approach to estimate the lower and the upper bounds for the Shattering coefficient function when tackling a specific supervised dataset from an analytical and numerical point of view, being more precise than Sauer-Shelah's approach~\cite{Sauer72jct,CIS-393795}. As a consequence, our contribution allows: (i) the estimation of the required number of hyperplanes in the best and worst-case scenarios, supporting the study, analysis and definition of learning architectures (e.g. the number of units in an artificial neural network); (ii) the estimation of the training sample sizes required to ensure supervised learning for practical applications; (iii) to address multi-class classification problems; (iv) considering multiple hyperplanes to derive the Shattering coefficient function; and (v) the comparison of dataset embeddings, once they (re)organize examples into some new space configuration. 

From a practical perspective, dataset embeddings and/or derived features can be assessed using our approach to conclude on the easier or more difficult space separability, thus leading to a space complexity measure. Finally, different space embeddings can be compared to one another in an attempt to select the most adequate to address some learning task. All those contributions together allow a more precise analysis on the space of admissible functions, a.k.a. the algorithm search bias $\mathcal{F}$, as well as the bias comparison against different learning settings. 

\ifCLASSOPTIONcompsoc
  \section*{Acknowledgments}
\else
  \section*{Acknowledgment}
\fi

I thank several people that somehow supported this work along several years throughout discussions and revisions, most specially Martha Dais Ferreira, Moacir Ponti, and Carlos Grossi. I also thank Thiago Bianchi for revising this paper.

This work was supported by Itaú Unibanco SA, São Paulo, Brazil; CNPq, Brazil, under the grant number 302077/2017-0; and the Center of Mathematical Sciences Applied to Industry (CEPID-CeMEAI) under grant 2013/07375-0. Any opinions, findings, and conclusions or recommendations expressed in this material are those of the authors and do not necessarily reflect the views of Itaú Unibanco SA, CNPq, nor FAPESP. 

\bibliographystyle{IEEEtran}


\begin{IEEEbiography}[{\includegraphics[width=1in,height=1.25in,clip,keepaspectratio]{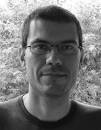}}]{Rodrigo Fernandes de Mello}
is currently Chief Data Scientist with Itaú Unibanco SA. His research interests include Statistical Learning, Machine Learning, and Data Streams.
\end{IEEEbiography}

\end{document}